\pdfoutput=1

\documentclass[11pt]{article}

\usepackage[]{acl}

\usepackage{times}
\usepackage{latexsym}

\usepackage[T1]{fontenc}

\usepackage[utf8]{inputenc}

\usepackage{microtype}

\DeclareFixedFont{\ttb}{T1}{txtt}{bx}{n}{8} 
\DeclareFixedFont{\ttm}{T1}{txtt}{m}{n}{8}  

\usepackage{color}
\definecolor{deepblue}{rgb}{0,0,0.5}
\definecolor{deepred}{rgb}{0.6,0,0}
\definecolor{deepgreen}{rgb}{0,0.5,0}

\usepackage{listings}

\newcommand\pythonstyle{\lstset{
    language=Python,
    basicstyle=\ttm,
    morekeywords={self},              
    keywordstyle=\ttb\color{deepblue},
    emph={MyClass,__init__},          
    emphstyle=\ttb\color{deepred},    
    stringstyle=\color{deepgreen},
    frame=tb,                         
    showstringspaces=false
}}

\lstnewenvironment{python}[1][]
{
\pythonstyle
\lstset{#1}
}
{}

\newcommand\pythonexternal[2][]{{
\pythonstyle
\lstinputlisting[#1]{#2}}}

\newcommand\pythoninline[1]{{\pythonstyle\lstinline!#1!}}

\title{Repro: An Open-Source Library for Improving the Reproducibility and Usability of Publicly Available Research Code}

\author{Daniel Deutsch and Dan Roth \\
  Department of Computer and Information Science \\
  University of Pennsylvania \\
  \texttt{\{ddeutsch,danroth\}@seas.upenn.edu}}

\newif\ifcomments
\commentstrue
\ifcomments
    \providecommand\dd[1]{\textcolor{blue}{[DD: #1]}}
    \providecommand\dr[1]{\textcolor{blue}{[DR: #1]}}
    \providecommand\todo[1]{\textcolor{red}{[TODO: #1]}}
\else
    \providecommand{\dd}[1]{}
    \providecommand{\dr}[1]{}
    \providecommand{\todo}[1]{}
\fi

\begin{document}
\maketitle

\begin{abstract}
We introduce Repro, an open-source library which aims at improving the reproducibility and usability of research code.
The library provides a lightweight Python API for running software released by researchers within Docker containers which contain the exact required runtime configuration and dependencies for the code.
Because the environment setup for each package is handled by Docker, users do not have to do any configuration themselves.
Once Repro is installed, users can run the code for the 30+ papers currently supported by the library.
We hope researchers see the value provided to others by including their research code in Repro and consider adding support for their own research code.\footnote{ \url{https://github.com/danieldeutsch/repro}
}
\end{abstract}
\section{Introduction}
Running the code released by the original authors of a research paper can be difficult.
There are often challenges in replicating the required runtime environment due to installing the correct versions of external libraries or putting resource files in the correct locations.
Further, the software packages may not have an easy-to-use API, making it hard to figure out how to use the released code.

In this work, we describe Repro, an open-source library which aims at improving the reproducibility and usability of publicly available research code.
Repro is a library of lightweight Python wrappers around code released by the authors of papers which provide a simple interface for running the original code without users needing to maintain or setup the necessary runtime environments themselves.

Repro achieves this using Docker, a platform for packaging together software applications along with all of the necessary dependencies.
Each of the papers supported in Repro has a corresponding Docker image that contains the exact runtime environment and dependencies for the code.
Then, Repro exposes a simple Python API for passing data to the Docker containers, processing the data with the original code, and returning the output to the user.

Since each codebase's dependencies are fully contained within the Docker images, users of Repro do not need to put any effort into setting up the correct environment to run code from a paper.
Once Repro is installed, users can easily run any of the code from the 30+ papers supported by the library.

This paper describes why Docker is an ideal platform for releasing reproducible code (\S\ref{sec:docker}), the details about how Repro is implemented (\S\ref{sec:repro}), a discussion of the best practices for ensuring reproducible code we have learned from developing the library (\S\ref{sec:best_practices}), and some of its limitations (\S\ref{sec:limitations}).
We hope that users see the value Repro provides and consider contributing Docker images and Python wrappers for their own papers' code.

\section{Docker as a Tool for Reproducibility}
\label{sec:docker}

\subsection{Background}
Repro is built on top of Docker.\footnote{\url{https://www.docker.com/}}
Docker is a tool for packaging together software applications into isolated, standalone environments that contain all of the dependencies necessary to run the applications.
The environments, called Docker images, can specify which operating system is used, which version of software libraries are installed, and can include data files.

Docker images are built using Dockerfiles.
Dockerfiles are text files with a specific syntax that contain a series of commands which are executed in sequence to build the image.
They begin with a base image, such as a specific version of Ubuntu, followed by commands that can install software libraries, copy local data files into the image, etc.

Once an image is built, it can be run as a Docker container, which is similar to a virtual machine.
The container allows the user to run software in or interact with the environment specified by the image from their own host machine.
However, any modifications made within the container do not persist when the container terminates;
Every time a container is created, it starts with a fresh environment defined by the image.

Importantly, Docker images and containers are platform independent.
If two different machines run the same container, the containers' environments will be identical up to differences in the machines' hardware.

Docker images can be easily distributed for others to use through an image registry server such as DockerHub.\footnote{\url{https://hub.docker.com/}}
Users can upload their image binaries to DockerHub that others can then download and run, analogous to how GitHub is used to distribute software code.
Thus, once a developer creates a Docker image, it is easy for others to replicate that exact runtime environment on their own machines.

\subsection{The Advantages of Docker for Reproducibility}

It is often challenging and time consuming to reproduce results from research papers.
Papers that publicly release code often include links to download pre-trained models or resources and written instructions for replicating the necessary runtime environment for their software.
However, it is not uncommon for the environments to be under-specified or not specified at all, for paths to resource files to be hard-coded based to locations on the author's machine, for the link to the pre-trained model or required resources to be broken, etc.
These problems can be exacerbated over time when information about the original environment configuration is forgotten or deleted entirely, making the code difficult to run.

Docker offers a solution to many of the common problems researchers encounter when they try to reproduce a result from a paper using resources released by the authors.
In addition to releasing code and dependencies, authors could also release Docker images along with their papers that is the exact environment necessary to run their code.
Other researchers would then be able to run the code without having to worry about exact library versions or the locations of pre-trained models since these details would be taken care of by the Docker image.
Then, if the images were stored on a public Docker image registry, the environment necessary to run the code associated with the paper would exist indefinitely.

Thus, the advantages of Docker make it an ideal platform for building a library focused on the reproducibility of research code.
\section{Repro}
\label{sec:repro}
Repro is a Python-based library that aims at improving the reproducibility and usability of research code released along with papers.
It provides easy-to-use Python APIs for running the original code released by the authors, which can be done from a single, lightweight Python environment.
Once Repro is installed, users can run the code from any paper supported by Repro without any additional setup effort.

\paragraph{Improving Reproducibility}
Every codebase supported by Repro is packaged into its own Docker image, which contains the original source code, runtime environment, and necessary dependencies.
Repro provides a lightweight Python wrapper around the Docker image which facilitates launching a Docker container, transferring data to the container, running the original code in the original environment, and returning the results to the user.
Because the paper code is run within a Docker container, that runtime environment will be the same for all users of the library.
Thus, the environment configuration which reproduces results from the original paper can be replicated for all Repro users, improving the reproducibility of the original research.

\paragraph{Improving Usability}
Because the papers' code runs in Docker images, users of Repro do not need to maintain the runtime environments or dependencies themselves since they are encapsulated within the images.
For instance, they do not have to create a Python environment specific to a codebase, install the software dependencies, or download any resource files, such as pre-trained models, to specific locations.
All of this is taken care of by the Docker images, and thus Repro makes running the original research code far easier than before.

The only environment the users need to maintain is the one which Repro is installed in.
However, since Repro's wrappers around the Docker images do not have difficult-to-install dependencies, Repro's required Python environment from which all of the supported papers' code can be run is very lightweight.

Figure~\ref{fig:bart_example} contains an example of how a user can easily run BART \citep{LLGGMLSZ20} to generate a summary of an input document.
The BART model corresponds to a Python class that provides a function to run inference and return the summary.
Behind the scenes, Repro launches the Docker container which contains the original code and models released by the BART authors, passes the input to the container, runs the original code in the container to produce the summary, and returns the result to the user in the original Python process.
All of this processing is hidden from the user, so BART's Python API looks like any other standard Python function.

The API for each paper depends on what the code does.
For instance, reference-based text generation metrics will accept a text to score and some references, question-answering models will take some input text and a question, etc.
We have tried to standardize the inputs and outputs formats for the same types of models so it will be easier to quickly run multiple papers' code on the same inputs.

\begin{figure}
    \centering
    \pythonexternal{figures/example_code/example.py}
    \caption{
        An example code snippet for generating a summary with BART \citep{LLGGMLSZ20}.
        \pythoninline{model.predict()} launches the Docker container that contains the runtime environment for BART, uses a pre-trained BART model to generate a summary of the input text, and returns the result as a string type to \pythoninline{summary}.
    }
    \label{fig:bart_example}
\end{figure}

\paragraph{Installation \& Running}
The library itself is lightweight and has minimal dependencies.
Our goal is to make it as easy-as-possible to install, which can be done by Python's pip package manager.
Running Repro requires a host machine with Docker installed.

\paragraph{Communication with Docker}
Exchanging data between the host machine's Python process and the Docker container is done via the machine's file system.
First, the Python process serializes the data which needs to be processed to a directory on the host machine.
Then, the process launches the Docker container and mounts that directory to the container, which gives the container the ability to read and write to the file system of the host machine.
The Python process executes a command within the container to process the data, and the container serializes the result to the same mounted directory and terminates.
The process then loads the results and returns them to the user.

While this communication with Docker may sound complex, it is entirely hidden from users of Repro.
The Python API uses normal Python types as inputs and outputs even though the data processing is largely done in Docker.
As such, it looks the same as a standard Python function.
Therefore, users do not need to know how to use Docker in order to use the library.

\paragraph{Distributing Docker Images}
All of the Docker images supported by Repro are hosted in DockerHub and have corresponding Dockerfiles in Repro.
When a new Dockerfile is added to Repro, a GitHub Action\footnote{\url{https://github.com/features/actions}} is triggered which builds an image from the Dockerfile and pushes it to DockerHub.

If a user attempts to run code in a Docker container for which the corresponding image is not present on their machine, Repro will automatically download that image for them.
The images can be manually downloaded from DockerHub as well.

\paragraph{Papers Implemented in Repro}
As of this writing, there is code from 30+ papers supported by Repro.
The majority of them are related to evaluating generated text based on our own research interests, but there are also models for text summarization, question-answering, question generation, constituency parsing, and more.
See Appendix~\ref{appendix:list_of_papers} for the full list.

Once a user installs Repro on a machine with Docker, all 30+ of the codebases can be run without any additional setup.
We are continually supporting more papers and welcome contributions by the research community.

\paragraph{Contributing}
We hope that the research community sees the benefits of Repro and contributes their own Docker images to the library for others to use.
Because many people within the NLP community may not have experience using Docker, our GitHub repository contains tutorials that explain how to install Docker, list basic Docker concepts and useful commands, and provide instructions for packaging a codebase into a Docker image.
We additionally explain how to write the Repro wrapper around the Docker image.

\section{Reproducibility Best Practices}
\label{sec:best_practices}
During the process of building Docker containers for various libraries, we have identified several best practices that future researchers can use to improve the ease-of-use of faithfully running their code as intended.

First, example inputs and expected outputs should be included in the code's documentation to help to ensure that reproductions are faithful to the original implementation.

The exact programming language environment should be specified in the documentation.
For example, this could include the version of Python as well as the versions of the Python packages that were used in the original environment.
In the case of Python, package managers such as pip and conda provide tools for exporting a list of installed packages to a file for distributing to others.

External resources, such as data files or pre-trained models, should be stored in locations which are not likely to be moved or deleted.
Anecdotally, we have found that files stored in locations owned by organizations (e.g., universities or companies) are more stable than those stored in individuals' personal storage platforms (e.g., Google Drive).

Authors should provide a command line interface for running their code end-to-end that accepts a file(s) as input and writes a file as output instead of a series of scripts which need to be run in series to process the data.
This makes it easier for other authors to run the code on their own data as well as integrate it into the Repro library because it makes the code easier to use.
\section{Hardware Limitations}
\label{sec:limitations}
Although Repro significantly improves the ease of reproducing the correct runtime environment and the usability of research code, the library has some limitations surrounding hardware compatibility.

Docker containers are only identical across machines up to differences in hardware.
As new hardware is released, it may not be compatible with older software, which could potentially make it difficult to run older Docker images.
For example, a new GPU may require software to use a minimum version of CUDA, and a model may use a specific version of PyTorch \citep{PGMLBCKLGADKYDRTCSFBC19} that is incompatible with any version of CUDA that can be used with the GPU.
Thus, the model will not run.
However, this issue can be mitigated by the authors updating their code and Docker images to be compatible with current hardware.

\section{Related Work}
There are various other software libraries which aim at making running a variety of research code easier.
The SacreBLEU \citep{Post18}, SacreROUGE \citep{DeutschRo20}, Huggingface Datasets \citep{LMJTPPCDPTDCMBeSSXPMSGDMDrBCGMLReW21}, and the GEM metrics\footnote{\url{https://github.com/GEM-benchmark/GEM-metrics}} libraries provide wrappers around or implementations of various text generation evaluation metrics in order to establish standardized implementations and make the metrics easier to run.
Libraries such as AllenNLP \citep{GGNTDLPSZ18} or Transformers \citep{WDrSCDMCRLFDSPMJPXSGDLRe20} provide frameworks for which researchers can train deep learning models.
Once someone is familiar with one of these frameworks, running a trained model which uses them is relatively straightforward since models within these frameworks typically use similar APIs.

They key difference between these approaches and Repro is how the environments for the models or metrics are maintained.
These libraries attempt to either have one complex runtime environment for all models/metrics they support or have a new environment for each one.
Repro instead has one lightweight Python environment for the codebase wrappers and uses Docker to maintain the paper-specific runtime environments, making the research code included in Repro very easy to use since the environments do not need to be maintained by the library users.

\section{Conclusion}
We have introduced Repro, a library built on Docker that aims at improving the reproducibility and usability of research code.
We hope that other researchers see how the library makes running their code more accessible to others and consider contributing their own Docker images.

\bibliography{bibliography}
\bibliographystyle{acl_natbib}

\clearpage
\appendix

\section{List of Supported Papers}
\label{appendix:list_of_papers}
The following is a list of papers with publicly available code that have implementations in Repro:
\begin{itemize}
    \item BART \citep{LLGGMLSZ20}
    \item BARTScore \citep{YuanNeLi21}
    \item BERTScore \citep{ZKWWA20}
    \item BaryScore \citep{CSCP21}
    \item BertSumExtAbs \citep{LiuLa19}
    \item BLANC \citep{VasilyevDhBo20}
    \item BLEU \citep{PRWZ02}
    \item BLEURT \citep{SellamDaPa20}
    \item CLIPScore \citep{HHFBC21}
    \item COMET \citep{RSFL20}
    \item DAE \citep{GoyalDu20}
    \item DepthScore \citep{SMCCd21}
    \item FactCC \citep{KMXS20}
    \item FEQA \citep{DurmusHeDi20}
    \item GSum \citep{DLHJN21}
    \item InfoLM \citep{ColomboClPi21}
    \item LERC \citep{CSSG20}
    \item Lite3Pyramid \citep{ZhangBa21}
    \item Meteor \citep{DenkowskiLa14}
    \item MoverScore \citep{ZPLGME19}
    \item NMN-Drop \citep{GLRSG20}
    \item NUBIA \citep{KKAAC20}
    \item Prism \citep{ThompsonPo20}
    \item QA\-Eval \citep{DeutschBeRo21}
    \item QuestEval \citep{SDLPSWG21}
    \item ROUGE \citep{Lin04}
    \item SummaQA \citep{SLPS19}
    \item SUPERT \citep{GaoZhEg20}
    \item the question generation model from \citet{PRMGTD21}
    \item the constituency parser from \citet{KitaevCaKl19}
    \item the recipe generator from \citet{DIKC20}
    \item the truecaser from \citet{SusantoChLu16}
    \item the QA-SRL parser from \citet{FMHZ18}
\end{itemize}

\end{document}